\documentclass{article}

\usepackage[final, nonatbib]{arxiv}

\RequirePackage{algorithm}
\usepackage{algorithmic}

\usepackage[utf8]{inputenc} 
\usepackage[T1]{fontenc}    
\usepackage{hyperref}       
\usepackage{url}            
\usepackage{booktabs}       
\usepackage{amsfonts}       
\usepackage{nicefrac}       
\usepackage{microtype}      

\usepackage{amsmath}
\usepackage{bm}
\usepackage{bbm}
\usepackage{xcolor}
\usepackage{graphicx}
\usepackage{overpic}
\usepackage{wrapfig}

\usepackage{inconsolata}

\usepackage{siunitx}
\newcommand{\cgrad}{\operatorname{grad}}
\DeclareMathOperator*{\cdiv}{div}
\DeclareMathOperator*{\dd}{d}
\DeclareMathOperator*{\diag}{diag}
\DeclareMathOperator*{\vect}{vec}

\newcommand{\code}[1]{\texttt{#1}}
\newtheorem{prop}{Proposition}
\newtheorem{thm}{Theorem}
\newtheorem{lem}{Lemma}

\usepackage[verbose=true,letterpaper]{geometry}
\AtBeginDocument{
	\newgeometry{
		textheight=8.8in,
		textwidth=6.3in,
		top=1in,
		headheight=14pt,
		headsep=25pt,
		footskip=30pt
	}
}
\title{A Differential Geometry Perspective on Orthogonal Recurrent Models}

\author{
    Omri Azencot \\
    Ben-Gurion University \\
    \texttt{azencot@cs.bgu.ac.il} \\
    \And
    N. Benjamin Erichson \\
    ICSI and UC Berkeley \\
    \texttt{erichson@berkeley.edu} \\
    \AND
    Mirela Ben-Chen \\
    Technion -- Israel Institute of Technology \\
    \texttt{mirela@cs.technion.ac.il} \\
    \And
    Michael W. Mahoney \\
    ICSI and UC Berkeley \\
    \texttt{mmahoney@stat.berkeley.edu} \\
}

\begin{document}

\maketitle

\begin{abstract}

Recently, orthogonal recurrent neural networks (RNNs) have emerged as state-of-the-art models for learning long-term dependencies. 
This class of models mitigates the exploding and vanishing gradients problem by design. 
In this work, we employ tools and insights from differential geometry to offer a novel perspective on orthogonal RNNs. 
We show that orthogonal RNNs may be viewed as optimizing in the space of divergence-free vector fields. 
Specifically, based on a well-known result in differential geometry that relates vector fields and linear operators, we prove that every divergence-free vector field is related to a skew-symmetric matrix. 
Motivated by this observation, we study a new recurrent model, which spans the entire space of vector fields. 
Our method parameterizes vector fields via the directional derivatives of scalar functions.
This requires the construction of latent inner product, gradient, and divergence operators. 
In comparison to state-of-the-art orthogonal RNNs, our approach achieves comparable or better results on a variety of benchmark tasks. 
  
\end{abstract}

\section{Introduction}

Recurrent Neural Networks (RNNs) are commonly used for modeling time-series data. Unfortunately, standard RNNs are challenging to train due to the exploding and vanishing gradient problem~\cite{bengio1994learning}. Recently, Arjovsky et al.~\cite{arjovsky2016unitary} showed that the dynamic behavior of gradients during training is fully determined by the eigenvalues of the hidden-to-hidden weight matrices. Their analysis led to the design of unitary weights RNNs, with several follow-up works \cite{wisdom2016full,jing2017tunable,helfrich2017orthogonal,mhammedi2017efficient,vorontsov2017orthogonality,chang2019antisymmetricrnn,maduranga2019complex,lezcano2019cheap,casado2019trivializations,kerg2019non}. From a mathematical viewpoint, restricting the weights to be unitary or orthogonal is equivalent to optimizing over the Stiefel manifold, which can be parameterized using skew-Hermitian matrices and the matrix exponential operation \cite{lezcano2019cheap,casado2019trivializations}. While unitary RNNs achieve state-of-the-art results on a few long-term memory tasks, their expressivity is \emph{limited}. This issue was partially-addressed by relaxing the unitary constraints~\cite{vorontsov2017orthogonality}, and through the design of non-normal connectivity weights~\cite{kerg2019non}.

In general, the class of sequence models including RNNs are known to be intimately related to the practice and theory of dynamical systems. 
Motivated by this connection, many approaches that incorporate dynamics properties into their models have been proposed. One example involves the learning of invariant quantities via their Hamiltonian or Lagrangian representations~\cite{lutter2019deep,greydanus2019hamiltonian,chen2019symplectic,zhong2019symplectic,toth2019hamiltonian}. Other techniques draw inspiration from Koopman theory, yielding neural network models where the evolution operator is linear~\cite{takeishi2017learning,morton2019deep,li2019learning,erichson2019physics,azencot2020forecasting}. Given the success of unitary RNNs in processing sequential data and their rich algebraic structure that is related to the Stiefel manifold, we raise the following interesting question:
\begin{quote}
	\emph{What is the {geometrical} interpretation of unitary RNNs, and how do we model and exploit it?}
\end{quote}

In this paper, we show that orthogonal RNN models may be viewed as optimizing in the space of \emph{divergence-free vector fields}. 
Based on a well-known result in differential geometry which relates vector fields and linear operators, we prove that every divergence-free vector field is related to a skew-symmetric matrix. These matrices are the generators of the orthogonal group via the exponential map \cite{lezcano2019cheap}. Our geometric perspective makes it clear that orthogonal units merely learn a \emph{single} vector field per layer, allowing one to exploit tools from differential geometry, ordinary differential equations, and dynamical systems for further analysis. Furthermore, we recall that vector fields with zero divergence span only a fraction of the space of all vector fields. Given these results, we propose a new recurrent model that we call the \emph{latent vector field recurrent neural network} (\code{lvfRNN}). Among the theoretical advantages of this model are that its hidden-to-hidden matrix represents an object with rich theoretical foundations and that it spans the whole space of vector fields.


To derive our method, we observe that vector fields can be parameterized via the \emph{directional derivative} of scalar functions (Sec.~\ref{subsec:divfree_skewsymm}). Namely, it is sufficient to know the directional derivative of all scalar functions with respect to a vector field to fully reconstruct it. The opposite is also true---given a vector field, it is easy to obtain its associated directional derivative. Building the latter operator requires the design of a differential geometry toolbox, including tools such as the gradient and divergence operators, as well as the inner product (Sec.~\ref{subsec:lvfrnn_mtd}). We then discuss several features of our approach with respect to the continuous setting, including its relation to unitary RNNs, and we offer simple criteria for training stable \code{lvfRNN} models (Sec.~\ref{subsec:lvfrnn_prop}). Finally, we describe certain implementation details (Sec.~\ref{sec:imp}), and we evaluate our model on several sequence tasks (Sec.~\ref{sec:results}).

Overall, the contributions of our work are the following.  
\begin{itemize}
	\item We propose a differential geometry interpretation of orthogonal RNNs, showing that these models optimize in the space of divergence-free vector fields. Furthermore, we prove that every skew-symmetric matrix is associated with a divergence-free vector field. 
	
	\item Our geometric perspective establishes the foundation to exploit tools from differential geometry, dynamical systems, and ordinary differential equations for further analysis and design of orthogonal recurrent models.
	
	\item We suggest a novel recurrent model, \code{lvfRNN}, whose span covers the entire space of vector fields, including those fields that are divergence-free. We show that our approach attains comparable or better results than state-of-the-art unitary RNN methods on several sequence benchmark tasks. 
\end{itemize}

\section{Related Work}
\label{sec:related}

\paragraph{Unitary and orthogonal RNN.}
 Arjovsky et al.~\cite{arjovsky2016unitary} proposed to constrain the recurrent weights to be the product of unitary matrices and Fourier transforms. Their approach achieved superior results in comparison to several other RNN baselines on problems that require long-term memory such as the copy and add tasks. However, their method can only represent a limited subspace of orthogonal matrices. To alleviate this issue, \cite{wisdom2016full}, and \cite{vorontsov2017orthogonality} used the Cayley transform on skew-symmetric matrices, and \cite{helfrich2017orthogonal}, and \cite{maduranga2019complex} further suggested to scale the latter transform to include the negative one eigenvalue. A complete parameterization of the orthogonal group with skew-Hermitian matrices was proposed in \cite{lezcano2019cheap,casado2019trivializations}. These latter works leverage concepts from Lie theory and the connection between a Lie algebra and its group via the exponential map. To further improve the expressivity of orthogonal transformations, Kerg et al.~\cite{kerg2019non} advocated the design of nonnormal hidden-to-hidden weights whose eigenvalues have unit norm. 

\paragraph{Dynamical systems and deep learning.}
Viewing deep learning through the lens of continuous-time dynamical systems has recently inspired several works.  
A large body of work focuses on constructing models by approximating continuous-time dynamical systems   \cite{chen2018neural,kidger2020neural,queiruga2020continuous}. Such models can also be studied by using tools from  numerical analysis \cite{lu2018beyond,yang2020dynamical,zhang2019towards}. Dynamical systems theory also provides useful tools for studying RNNs \cite{vogt2020lyapunov,engelken2020lyapunov,chang2019antisymmetricrnn,erichson2020lipschitz,lim2021noisy}.

\paragraph{Physics-based models.}
Incorporating physics priors into neural networks is a natural choice when processing time-series data. A particular interest was given to the automatic learning of equivariances in dynamical systems through their Hamiltonian~\cite{greydanus2019hamiltonian,chen2019symplectic,zhong2019symplectic,toth2019hamiltonian} or Lagrangian~\cite{lutter2019deep} formulations. Another successful ansatz models the nonlinear dynamics using Koopman theory, where it is assumed that the inputs can be embedded in a way such that their evolution is linear \cite{takeishi2017learning}. In this context, the infinitesimal generator of the Koopman operator is closely-related to latent vector fields and to the theory that underlies our methodology~\cite{lasota2013chaos}. However, to the best of our knowledge, the relation between vector fields and recurrent neural networks is novel to this work. More fundamentally, our approach takes a further step in the direction of combining machine learning and differential geometry knowledge. The \code{lvfRNN} model establishes the link between recurrent networks and their geometrical interpretation while laying the groundwork for further development using differential geometry~\cite{spivak1970comprehensive} and differential equations~\cite{lang2012fundamentals} literature.

\paragraph{Vector fields and machine learning.}
Vector fields are frequently used in the literature of machine learning across various inference tasks. For instance, \cite{lin2012multi} improve multi-task learning using vector fields by characterizing the differential structure of tasks while exploiting the geometric structure of the data. In~\cite{perrault2011directed}, the authors recover an Euclidean embedding of a weighted directed graph by associating it with a manifold and a vector field along with Laplacian-type operators. More recently, \cite{cohen2019general} develop a general theory of group equivariant convolutional neural networks (G-CNN) by showing that convolution kernels are equivalent to linear maps, allowing to analyze G-CNN via the theory of fiber bundles. Vector fields also appear in generative adversarial networks to improve training by promoting the update steps to be conservative~\cite{mescheder2017numerics} or to visualize the optimization landscape~\cite{berard2019closer}. Another common task of deep learning is the solution of differential equations by modeling the space of their admissible vector fields, see, e.g.,~\cite{trischler2016synthesis,chen2018neural,berg2018unified}. Perhaps closest in spirit to our work is the interpretation of recurrent models as dynamical systems~\cite{amari1972characteristics,hopfield1984neurons,tsung1995phase} where the map between hidden states may be viewed as a vector field. However, our point of view is fundamentally different from this classical perspective on recurrent models, as we link vector fields to matrices through their action on scalar functions. Given the current literature in the field, we believe that our work offers a novel complementary interpretation of sequential models.

\section{A Differential Geometry Perspective}
\label{sec:lvfrnn}

In this section, we briefly review a neural model which optimizes over finite-dimensional groups, as was recently suggested in \cite{lezcano2019cheap}. We then consider a differential geometry perspective which extends this model to \emph{infinite-dimensional} Lie groups of tangent vector fields. Finally, we discuss some of the inherent challenges involved in accomplishing a practical extension that respects the underlying geometric structure.

\subsection{Neural models over matrix groups}

The baseline model we consider was proposed to mitigate the issue of exploding and vanishing gradients, and to better span the space of orthogonal connectivity matrices. Given an input vector $x_t \in \mathbb{R}^m$, the \code{expRNN} model~\cite{lezcano2019cheap} describes the evolution of the hidden state $h_t \in \mathbb{R}^\kappa$ by
\begin{align} \label{eq:exprnn}
	h_t = \sigma(\exp(A)h_{t-1} + U x_t) \ ,
\end{align}
where $\sigma(\cdot)$ is some nonlinear activation, $U \in \mathbb{R}^{\kappa \times m}$ embeds the input onto the latent space, and $\exp(A)$ is the matrix exponential that maps the skew-symmetric $A \in \mathrm{Skew}(\kappa)$ onto an orthogonal matrix. We note that other models such as the unitary RNN (\code{uRNN})~\cite{arjovsky2016unitary}, full capacity unitary RNN (\code{fcuRNN})~\cite{wisdom2016full}, scaled Cayley orthogonal RNN (\code{scoRNN})~\cite{helfrich2017orthogonal}, and other related approaches are encapsulated in the above model~\eqref{eq:exprnn}. 

The unitary RNN methods share the key advantage that their hidden-to-hidden matrix is isometric, i.e., it preserves the norm of the vectors it acts on (gradients included). However, previous work considers the problem of designing isometric maps strictly from an \emph{algebraic} viewpoint---exploiting the relation between skew-symmetric and orthogonal matrices. In what follows, we advocate that the \emph{differential geometry} viewpoint is advantageous and it should also be taken into account. To this end, we observe that unitary RNN can be viewed as optimizing in the space of divergence-free vector fields (\ref{subsec:divfree_skewsymm}), and we propose a new recurrent model which uses the entire space of vector fields (\ref{subsec:lvfrnn_mtd}). Specifically, we provide a concrete discrete differential geometry toolbox that respects many of the continuous properties (\ref{subsec:disc_diff_geom}) while allowing to easily control high-level features such as dynamical stability and isometry (\ref{subsec:lvfrnn_prop}). 

\begin{figure*} [t]
	\centering
	\begin{overpic}[width=.85\textwidth]{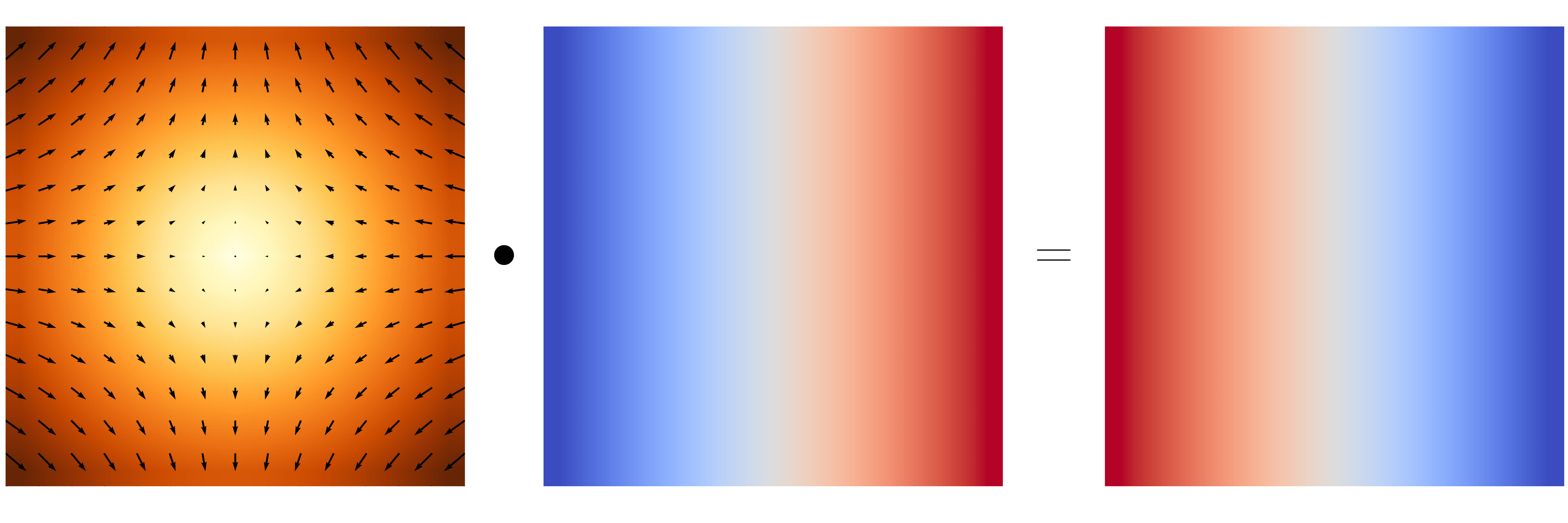} 
		\put(7,-1){$\mathfrak{v}=(-x,y)$} \put(45,-1){$\mathfrak{f}=x$} \put(77,-1){$\mathfrak{v}(\mathfrak{f})=-x$}
	\end{overpic}
	\caption{Vector fields may be viewed as \emph{operators} acting on scalar functions. For instance, applying the two-dimensional divergence-free vector field $\mathfrak{v}$ (left) on the scalar function $\mathfrak{f}$ (middle) is equivalent to computing the directional derivative of $\mathfrak{f}$ with respect to $\mathfrak{v}$. The resulting function is the $x$-coordinate of $\mathfrak{v}$ which is given by $\mathfrak{v}(\mathfrak{f})=-x$ (right).}
	\label{fig:vf_example}
\end{figure*}

\subsection{Divergence-free vector fields and skew-symmetric matrices}
\label{subsec:divfree_skewsymm}

In the continuous setting, a vector field $\mathfrak{v}$ can be thought of as a smooth assignment of a vector (direction and length) per point of the domain. These objects are indispensable in the study of natural phenomena, and the differentiation and integration of vector fields is directly associated to their features. For instance, the \emph{divergence} of a vector field $\nabla \cdot \mathfrak{v}$ captures the amount of outward flux per point, and it is formally defined as the trace of the Jacobian of $\mathfrak{v}$. We say that a vector field is divergence-free if at every point it has zero divergence. 

A classical approach in differential geometry views vector fields as operators which act on scalar functions~\cite{frankel2011geometry}. Namely, given a continuously differentiable function $\mathfrak{f}$, the action of the vector field $\mathfrak{v}$ on $\mathfrak{f}$ is given by
\begin{align} \label{eq:vf_op}
	\mathfrak{v}(\mathfrak{f}) = \langle \mathfrak{v}, \nabla \mathfrak{f} \rangle := \mathcal{D}_\mathfrak{v} (\mathfrak{f}) \ ,
\end{align}
where $\nabla$ is the gradient operation and $\langle \cdot, \cdot \rangle$ is a pointwise inner product. From a geometric point-of-view, $\mathcal{D}_\mathfrak{v} (\mathfrak{f})$ is the directional derivative of $\mathfrak{f}$ in $\mathfrak{v}$'s direction. A basic example in 2D of a divergence-free vector field and its action on a particular function is shown in Fig.~\ref{fig:vf_example}. Specifically, we consider $\mathfrak{v} = (-x,y)$ and $\mathfrak{f} = x$. Clearly, $\nabla\cdot \mathfrak{v} = 0$ and $\mathfrak{v}(\mathfrak{f}) = \langle \mathfrak{v}, \nabla \mathfrak{f} \rangle = -x$. 

Eq.~\eqref{eq:vf_op} shows that $\mathcal{D}_\mathfrak{v}$ is a \emph{linear} operator that is independent of its input. Importantly, directional derivative operators are infinite-dimensional, even if the underlying domain is finite-dimensional, e.g., $\mathbb{R}^m$ in our case. Interestingly, $\mathcal{D}_\mathfrak{v}$ fully encodes $\mathfrak{v}$, and thus given a vector field, we can construct its related $\mathcal{D}_\mathfrak{v}$ and vice versa. For instance, the example in Fig.~\ref{fig:vf_example} highlights that a vector field can be reconstructed from its directional derivative by applying it to the coordinate functions of the domain. The following result establishes the relation between divergence-free vector fields and skew-symmetric (infinite-dimensional) matrices. 
\begin{thm} \label{thm:divfree_skew}
	The vector field $\mathfrak{v}$ is divergence-free if and only if its operator $\mathcal{D}_\mathfrak{v}$ is skew-symmetric.
\end{thm}

\paragraph{Sketch of the proof.} The key observation in the detailed proof we provide in App.~\ref{app:divfree_skew} is that the operator $\mathcal{D}_\mathfrak{v}$ acts on scalar functions which can be represented with a basis $\{ \phi_j \}$. Then, we can compute the individual elements of $\mathcal{D}_\mathfrak{v}$ by using the standard inner product for functions $\mathfrak{f}, \mathfrak{g}$ in the $L^2$ function space, which is given by
\[
    \langle \mathfrak{f},\, \mathfrak{g} \rangle_\mathcal{M} := \int_\mathcal{M} \mathfrak{f}(x) \mathfrak{g}(x) \dd x \ .
\]
From a direct calculation and the use of standard vector calculus identities, it follows that
\begin{align*}
    (\mathcal{D}_\mathfrak{v})_{ij} &= \langle \phi_i,\, \mathcal{D}_\mathfrak{v}(\phi_j) \rangle_\mathcal{M} \\
    &= - \langle \mathcal{D}_\mathfrak{v}(\phi_i),\, \phi_j \rangle_\mathcal{M} = - (\mathcal{D}_\mathfrak{v})_{ji} \ .
\end{align*}
The second equality holds on domains with no boundary and vector fields with zero divergence as we fully explain in the proof in App.~\ref{app:divfree_skew}. Importantly, this theorem allows one to view orthogonal RNNs as vector field networks in the \emph{limit}. Thus, it motivates one to exploit tools and results associated with vector fields to design and improve neural network models, as we show in what follows.

\subsection{A latent vector field perspective}
\label{subsec:lvfrnn_mtd}

While Thm.~\ref{thm:divfree_skew} characterizes isometric recurrent models such as \code{expRNN} using vector fields, it also naturally raises the following question: What about fields whose divergence is nonzero? Indeed, the subset of divergence-free vector fields represents only a fraction of the whole space. Thus, a recurrent model that spans all possible vector fields would be more expressive in comparison to orthogonal networks. In the smooth setting, the set of vector field operators forms a Lie (sub)group~\cite{marsden2013introduction}, and thus we would like to extend \code{expRNN} from the special orthogonal group $\mathrm{SO}(\kappa)$ to the setting of vector field objects.

\begin{figure*} [t]
	\centering
	\begin{overpic}[width=0.99\textwidth]{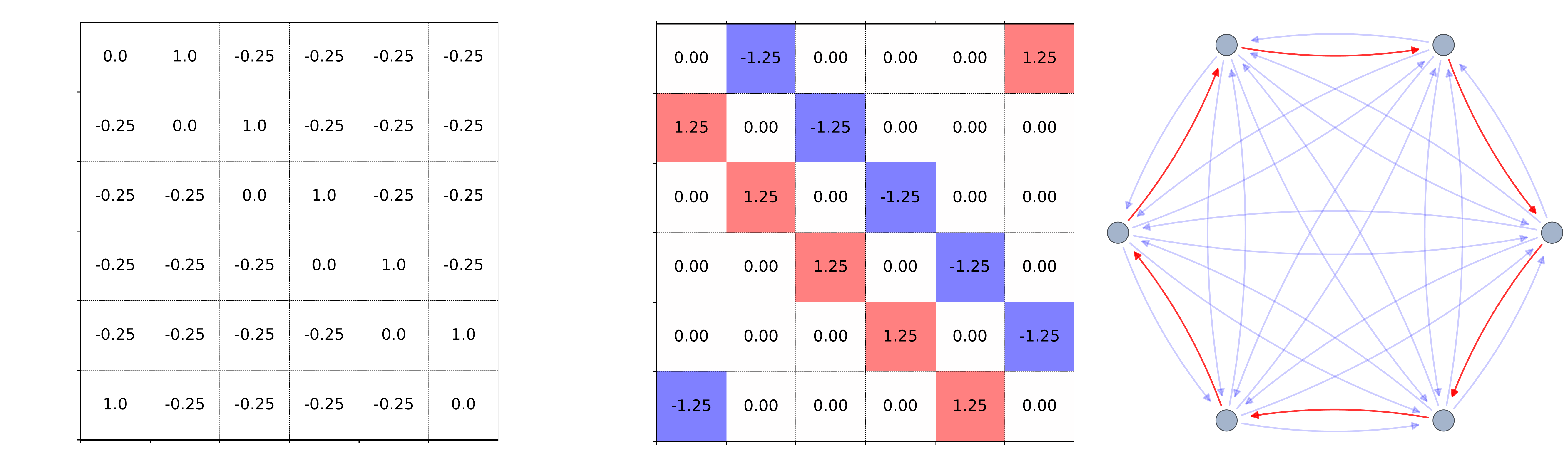} 
		\put(0,14.1) {$V =$} \put(35,14.1) {$\mathcal{D}_V =$}
	\end{overpic}
	\caption{We plot an example latent divergence-free vector field $V$ for $\kappa=6$ (left) with its associated directional derivative matrix $\mathcal{D}_V$ (middle). Viewing the latent domain as a complete graph, we visualize the vector field $V$ as prescribing directed weights between every pair of nodes, where in this example red corresponds to the value $1.0$ and blue is related to $-0.25$ values (right).}
	\label{fig:vf_sketch}
\end{figure*}

We consider the space of vector fields to be any $\kappa\times\kappa$ real-valued matrix with zero diagonal, i.e., $\mathrm{vf}_\kappa := \{ V \in \mathbb{R}^{\kappa \times \kappa} | \diag(V) =0 \}$. That is, $V \in \mathrm{vf}_\kappa$ encodes a vector field by including its $\kappa$-vectors in rows (see Fig.~\ref{fig:vf_sketch}, left). The diagonal of $V$ is set to zero to exclude vector fields with self-loops. The set $\mathrm{VF}_\kappa = \{ \mathcal{D}_V | V \in \mathrm{vf}_\kappa \}$ denotes the associated set of directional derivative operators. Given a specific realization of $\mathrm{VF}_\kappa$, a ``straightforward'' approach to extend \code{expRNN} would be to consider the following model
\begin{align} \label{eq:exp_lvfrnn}
	h_t = \sigma(\exp(\mathcal{D}_V) h_{t-1} + U x_t) \ , 
\end{align}
subject to $\mathcal{D}_V \in \mathrm{VF}_\kappa$, i.e., we restrict the matrix $\mathcal{D}_V$ to the class of vector field matrices of dimension $\kappa$, $\mathrm{VF}_\kappa$.

Unfortunately, there are two key differences between $\mathrm{SO}(\kappa)$ and vector fields which limit the validity of Eq.~\eqref{eq:exp_lvfrnn} here. First, the Lie group of the vector field is infinite-dimensional, and thus an approximation is required for~\eqref{eq:exp_lvfrnn}. Second, while the Lie bracket is closed for smooth directional derivative operators, it is not closed for the finite $\mathrm{VF}_\kappa$ as we show below. Thus, the matrix exponential of $\mathcal{D}_V$ does not optimize in $\mathrm{VF}_\kappa$ in the sense that $\frac{\partial}{\partial V}\exp(\mathcal{D}_V) \not\in \mathrm{VF}_\kappa$ for all $V \in \mathrm{vf}_\kappa$. Moreover, the next result, which we will prove in App.~\ref{app:lie_algebra}, emphasizes some of the difficulties in constructing a matrix vector fields group which respects Thm~\ref{thm:divfree_skew}.
\begin{thm}
	There is no finite-dimension matrix Lie algebra with elements $R + T$ where $R$ is skew-symmetric and $T$ is diagonal under the standard commutator, whose skew-symmetric matrices are associated with divergence-free vector fields.
\end{thm}

\paragraph{Sketch of the proof.} The main idea behind the proof is to show that a set of vector field operators $\mathrm{VF}_\kappa$ whose elements are sums of skew-symmetric and diagonal matrices, is not closed under the \emph{commutator} action, and thus it is not a Lie algebra. For matrix groups such as the one we consider, the standard commutator reads
\[
    [\mathcal{D}_U, \mathcal{D}_V] = \mathcal{D}_U \mathcal{D}_V - \mathcal{D}_V \mathcal{D}_U \ .
\]
A straightforward calculation shows that for general vector fields $U, V \in \mathrm{vf}_\kappa$, the commutator action yields a matrix whose diagonal is zero. However, we show in the detailed proof in App.~\ref{app:lie_algebra} that the resulting operator is not skew-symmetric, as it includes a \emph{symmetric} component which is not trivial in the general case. Thus, the commutator breaks the relation between divergence-free vector fields and skew-symmetric operators.

\section{The Latent Vector Field Model}

To address some of the challenges we mentioned above and to optimize in the space $\mathrm{VF}_\kappa$, we propose an approach that is based on numerical integrators of differential equations. Specifically, given a smooth vector field $\mathfrak{v}$, its associated transport equation describes the temporal evolution of a scalar function along the flow lines of $\mathfrak{v}$, and it is given by
\begin{align} \label{eq:transport_eq}
	\partial_t \mathfrak{f} = - \mathcal{D}_\mathfrak{v} \mathfrak{f} \ , 
\end{align}
for some initial condition $\mathfrak{f}(0) = \mathfrak{f}_0$. The above Eq.~\eqref{eq:transport_eq} is a linear ordinary differential equation, and it may be integrated with a variety of approaches. In this work, we focus on the explicit Euler integrator which in operator notation reads
\begin{align} \label{eq:euler_step}
	\mathcal{C}_V &= I - \tau \mathcal{D}_V \ , \quad \mathrm{s.t.} \;\; V \in \mathrm{vf}_\kappa \ , 
\end{align}
where $I$ is an identity matrix of size $\kappa$, and $\tau \in \mathbb{R^+}$ is the timestep. Thus, our approach can be viewed as optimizing in $\mathrm{VF}_\kappa$ by projecting onto that space at each and every iteration. Finally, we define the latent vector field recurrent neural network (\code{lvfRNN}) model via
\begin{align} \label{eq:lvfrnn}
	h_t = \sigma(\mathcal{C}_V h_{t-1} + U x_t) \ , \quad \mathrm{s.t.} \;\; \mathcal{D}_V \in \mathrm{VF}_\kappa \ .
\end{align}

To complete our construction, we need a proper definition of vector fields and their directional derivatives in a latent domain composed of $\kappa$ nodes. For simplicity, we assume that the underlying domain is fully connected with no self-edges, and thus information can propagate from any node to every other different node. Therefore, a latent vector field is a matrix $V \in \mathrm{vf}_\kappa \subset \mathbb{R}^{\kappa \times \kappa}$ with zero main diagonal. Additionally, we define a differential geometry toolbox, including an inner product and the gradient and divergence operators. Here, we briefly list these definitions, and we provide a more complete discussion and motivation for our choices in App.~\ref{app:diff_geom}. To avoid confusion with the continuous setting, we denote by $\cgrad$ and $\cdiv$ the discrete gradient and divergence operators, respectively. We show in Fig.~\ref{fig:vf_sketch} an example of a latent vector field, its associated $\mathcal{D}_V$ matrix, and its visualization as a weighted directed graph.

\subsection{A latent differential geometry toolbox}
\label{subsec:disc_diff_geom}

\paragraph{Gradient of a hidden state $h$.} In our setup, the hidden state $h \in \mathbb{R}^{\kappa}$ takes the role of a scalar function, and its gradient vector field $\cgrad h \in \mathbb{R}^{\kappa \times \kappa}$ per node is taken as the (forward) finite differences between the current node and all other nodes. Formally,
\begin{align} \label{eq:grad}
	(\cgrad h)_{ij} = h_i - h_j \ , \quad i,j = 1,2, ..., \kappa \ .
\end{align}

\paragraph{Divergence of a vector field $V$.} The divergence $\cdiv V \in \mathbb{R}^\kappa$ of a vector field is defined to be the sum of contributions of all incoming and outgoing edges related to node $i$, i.e.,
\begin{align} \label{eq:div}
	(\cdiv V)_i = \sum_{j=1}^\kappa V_{ji} - V_{ij} \ , \quad i = 1,2, ..., \kappa \ .
\end{align}

\paragraph{Directional derivative of $V$.} To encode the directional derivative action in Eq.~\eqref{eq:vf_op} for any hidden state, we propose to construct $\mathcal{D}_V \in \mathbb{R}^{\kappa \times \kappa}$ as follows,
\begin{align} \label{eq:dv_op}
	(\mathcal{D}_V)_{ij} =  \begin{cases}
		V_{ji}-V_{ij}       & \text{if $i \neq j$} \\
		-(\cdiv V)_i          & \text{otherwise}
	\end{cases}  .
\end{align}
It follows that any such operator can be decomposed into the sum of a diagonal part (flux) and a skew-symmetric part (rotation). Namely, for every $V$ we have that 
\begin{align}
    \mathcal{D}_V = R_V - T_V \ ,    
\end{align} 
where $R_V = V^T - V$ is a skew-symmetric matrix, i.e., $R_V + R_V^T = 0$, and $T_V$ is the diagonal matrix whose entries along the main diagonal satsify $T_V = \diag(\cdiv V)$.

\subsection{Properties of latent vector field models}
\label{subsec:lvfrnn_prop}

Our approach features several interesting properties which we discuss next. We show that $\mathrm{VF}_\kappa$ forms a vector space, allowing to design architectures that use linear combinations of different latent vector fields while staying in $\mathrm{VF}_\kappa$. We then position our model with respect to existing recurrent models regarding characterizing features such as the normality of $\mathcal{D}_V$ and whether its spectrum is imaginary. Finally, we show a simple relation between the divergence of the latent vector field and the stability of the dynamics in the sense of non-exploding trajectories.

\paragraph{Vector space structure and complexity measures.} The set of matrices $\mathrm{VF}_\kappa := \{ \mathcal{D}_V | V \in \mathrm{vf}_\kappa \}$ forms a vector space with the usual matrix addition and product with a scalar. Furthermore, we have that $\dim(\mathrm{VF}_\kappa) = \kappa(\kappa - 1)/2$ as the values of the skew-symmetric $R_V$ are arbitrary and they fix the diagonal of $T_V$. In terms of computational requirements, a na\"ive implementation of \code{lvfRNN} uses $\mathcal{O}(\kappa^2)$ space and $\mathcal{O}(\kappa^2)$ time, similar to vanilla {RNN} models. 

\paragraph{Relation to other recurrent models.} Most recurrent models whose hidden-to-hidden weight matrix has unit norm eigenvalues consider orthogonal or unitary matrices \cite{lezcano2019cheap} and their nonnormal extensions~\cite{kerg2019non}. In our setting, the vector fields space $\mathrm{VF}_\kappa$ includes normal and nonnormal matrices as well as operators whose spectrum is complex or imaginary. We provide a detailed classification in the next proposition where we show that for divergence-free latent fields, our model coincides with the matrix subspace of \code{expRNN}~\cite{lezcano2019cheap}. The proof of this result is based on basic linear algebra considerations and the unique structure of $\mathcal{D}_V$, and it is given in App.~\ref{app:lvfrnn_prop}.
\begin{prop}
	Let $\mathcal{D}_V \in \mathrm{VF}_\kappa$. Then $\mathcal{D}_V$ is normal and has imaginary spectrum if $(\cdiv V) = 0$. 
\end{prop}

\paragraph{Dynamical stability features.} The stability of recurrent neural networks plays a key role in their learning capabilities and training ease~\cite{miller2018stable,chang2019antisymmetricrnn,erichson2020lipschitz}. In particular, a non-stable system leads to an arbitrary growth of gradients, i.e., the exploding gradients problem. While in practice the gradients can be clipped when reaching a certain threshold~\cite{pascanu2012understanding}, we provide a simple characterization for the stability of \code{lvfRNN}. From a dynamical systems perspective, $\mathcal{D}_V$ matrices are stable if the real part of their eigenvalues is non-positive. The following result describes how the spectrum of $\mathcal{D}_V$ can be designed through the divergence of $V$. We prove the next proposition in App.~\ref{app:lvfrnn_prop}.
\begin{prop}
	Let $\mathcal{D}_V \in \mathrm{VF}_\kappa$. Then, the matrix $\mathcal{C}_V$ is stable if $(\cdiv V)_i \leq 0$ for every node $i=1,2,...,\kappa$.
\end{prop}

\subsection{Implementation}
\label{sec:imp}

The main difference between \code{lvfRNN} and a standard RNN is that the recurrent  matrix $\mathcal{C}_V$ is parameterized by the space of directional derivative matrices, as defined in~\ref{subsec:lvfrnn_mtd}. Thus, the learnable parameters of this layer depend on the latent vector field variable $V$ which is a $\kappa\times\kappa$ matrix excluding the main diagonal. We note that our model~\eqref{eq:lvfrnn} does not place any particular constraints on $\mathcal{D}_V$, e.g., with respect to its skew-symmetry. However, if long-term memory is required as in the copy task, one may augment the \code{lvfRNN} with a hard or soft constraint on the divergence of $V$. Specifically, to obtain a skew-symmetric $\mathcal{D}_V$ matrix, the additional constraint takes the form of $(\cdiv V)_i = 0$ for every node $i$, see also Thm.~\ref{thm:divfree_skew}. For example, as a soft penalty, the latter constraint can be implemented via $\lambda | \cdiv(V) |^2, \lambda \in \mathbb{R}^+$, which we add to the loss during training. We provide a baseline pseudocode implementation for the construction of $\mathcal{C}_V$ in Alg.~\ref{alg:lvfrnn_alg}.

\begin{algorithm}[hb]
    \caption{Construction of $\mathcal{C}_V$ matrices}
    \label{alg:lvfrnn_alg}
    \begin{algorithmic}[1]
        \STATE Input: matrix dimension $\kappa \in \mathbb{R}^+$, timestep $\tau \in \mathbb{R}^+$
	    \vspace{1mm}
		
		\STATE Initialize $V$ to be doubly-stochastic
		\vspace{1mm}
		
		\STATE $R_V = V^T - V$ \hspace{3cm} \# rotation component
		\vspace{1mm}
		
		\STATE $T_V = \diag(\mathrm{rows\_sum}(R_V))$ \hspace{1.05cm} \# flux component
		\vspace{1mm}
		
		\STATE $\mathcal{C}_V = I_\kappa -\tau(R_V - T_V)$
		\vspace{1mm}
		
		\STATE Return: $\mathcal{C}_V \in \mathbb{R}^{\kappa \times \kappa}$ following Eq.~\eqref{eq:euler_step}
    \end{algorithmic}
\end{algorithm}

To initialize $V$, we experimented with common choices such as Cayley~\cite{helfrich2017orthogonal} and Henaff~\cite{henaff2016recurrent} initialization schemes.  However, we noticed that the following method yields the best results. We propose to sample $V$ from the space of doubly stochastic matrices of size $\kappa$. While a true uniform sampler is challenging to achieve, the following scheme is easy to code and it provides reasonable results. We start with a $V_0 \in \mathbb{R}^{\kappa\times\kappa}$ sampled uniformly from $[0, 1]$, and we iteratively project it onto the space of row-stochastic (rows sum to one) and column-stochastic (columns sum to one) subspaces~\cite{sinkhorn1964relationship}. This procedure converges in a few iterations to an error threshold of $1\mathrm{e}{-8}$, producing a doubly-stochastic matrix $V$, see Alg.~\ref{alg:dsalg}. In our experiments, we evaluated the baseline unitary RNNs using our scheme as well as the Cayley and Henaff schemes. However, in most cases, we did not see any improvement when using doubly stochastic initial weights for unitary RNN models.

\begin{algorithm}[t]
    \caption{Random Sampling of Doubly Stochastic Matrices}
    \label{alg:dsalg}

    \begin{algorithmic}[1]

        \STATE{Input: matrix size $\kappa \in \mathbb{R}^+$, and threshold $\epsilon \in \mathbb{R}^+$}
        \vspace{1mm}

        \STATE{Output: $V \in \mathbb{R}^{\kappa \times \kappa}$ such that $V$ is doubly-stochastic}
        \vspace{1mm}
        
        \STATE{Initialize $V^0 = \mathrm{rand}({\kappa})$}    
        \vspace{1mm}
        
        \FOR{$l=0,1,2,...$}
        \vspace{1mm}
        
        \STATE{$V^l =  \diag(V^l \, 1_k)^{-1} V^l$}               
        \vspace{.5mm}
        
        \STATE{$V^{l+1} =  V^l \diag(1_k^T V^l)^{-1}$}               
        \vspace{1mm}
        
        \IF{$|| V^l 1_k - 1_k ||^2 + || 1_k^T V^{l+1}  - 1_k^T ||^2 < \epsilon$} 
            \STATE{break} 
        \ENDIF
        \vspace{1mm}
        
        \ENDFOR
        \vspace{1mm}
        \STATE{Return $V = V^{l+1}$}

    \end{algorithmic}
    
\end{algorithm}

\section{Experiments}
\label{sec:results}

In this section, we evaluate the performance of our approach and compare it to state-of-the-art unitary recurrent models such as \code{uRNN}~\cite{arjovsky2016unitary}, \code{euRNN}~\cite{jing2017tunable}, \code{fcuRNN}~\cite{wisdom2016full}, \code{expRNN}~\cite{lezcano2019cheap}, \code{nnRNN}~\cite{kerg2019non}, and \code{RNN}~\cite{rumelhart1986learning}. We focus on three learning tasks that are commonly used for benchmarking, the copy task~\cite{hochreiter1997long}, the polyphonic music task on the JSB and MuseData datasets \cite{allan2005harmonising,boulangerlew2012modeling}, the TIMIT speech prediction problem~\cite{halberstadt1999heterogeneous}, and the character level prediction task on the PTB dataset~\cite{marcus1993building}. We chose to focus on these tasks as they require from the modeling architecture long-term memory capabilities and relatively high expressivity. In what follows, we describe the results for the character-level, speech, and polyphonic music tasks, and we provide additional results in App.~\ref{app:results}.

In our experimental setup, we match the number of hidden units across all architectures for the copy and character-level prediction tasks with 182 and 1024 units, respectively. On the polyphonic music and speech prediction problems, we match the number of weights to $\approx 350$k and $\approx 200$k, respectively. For each of the tasks and for each model, we provide the loss over the test set, and we additionally report the accuracy for the language and copy tasks. To obtain these results, we performed an extensive hyperparameter search for each model based on the guidelines and code provided by the authors, and we list the hyperparameters that were used in practice in App.~\ref{app:hyperparam}. Overall, our method achieves good results on long-term memory tasks such as the copy task, whereas on tasks that require both high expressivity and memory, our approach attains the best results in comparison to other approaches.

\subsection{Character level prediction task} 

This language task involves a long string of characters which is split using truncated backpropagation through time, and the goal is to predict the next character for every position of the sub-strings. In our experiments, we compare the different models for a corpus split of $150$ time steps. The loss function for this task is the cross entropy, and the performance is measured using the mean bits per character (BPC). We used the Penn Tree Bank (PTB) dataset which contains $5059$k characters for training, $396$k for validation, and $446$k for testing, with an alphabet size of $50$. For instance, here is a short excerpt from the train set
\begin{center}
	\texttt{... t h e \_ a v e r a g e \_ s e v e n - d a y \_ c o m p o u n d \_ ...}
\end{center}
We show in Tab.~\ref{tab:ptb_res} the loss and accuracy results we obtained for this task. These results highlight the expressivity of our model since the hidden size is fixed in this setting. For comparison, \code{nnRNN} uses $\approx 2.8$m trainable parameters, whereas our method exploits only $\approx 1.3$m parameters (similar to \code{expRNN} and \code{fcuRNN}).

\begin{table}[!h]
    \caption{Using a fixed amount of $1024$ hidden units, we evaluate the performance of various models on PTB and list their obtained BPC and accuracy values. (Lower values for BPC are better.)}
    \vspace{+0.3cm}
    \centering
    \begin{tabular}{cccc}
        \toprule
        Method & BPC & Accuracy \\
        \midrule
        \code{RNN}                                  & $1.65$        & $62.23\%$     \\
        \code{uRNN}~\cite{arjovsky2016unitary}    & $1.62$        & $65.81\%$     \\
        \code{euRNN}~\cite{jing2017tunable}       & $1.61$        & $65.68\%$     \\
        \code{fcuRNN}~\cite{wisdom2016full}       & $1.50$        & $68.01\%$     \\
        \code{expRNN}~\cite{lezcano2019cheap}     & $1.49$        & $68.07\%$     \\ 
        \code{nnRNN}~\cite{kerg2019non}           & $1.47$        & $68.78\%$     \\
        Ours                                        & \bm{$1.43$}   & \bm{$69\%$}   \\
        \bottomrule 
    \end{tabular} \label{tab:ptb_res}
\end{table}

\subsection{TIMIT speech task}

The goal in the speech data task is to predict the log-magnitude of real-world speech frames of a short-time Fourier transform (STFT). In the spectral domain, an audio signal is encoded using a complex-valued matrix of size $F \times T$, where $T$ is the number of frames and $F$ represents the number of frequency bins. Reconstructing the audio signal is achieved by using the phase of the original inputs. Given the log-magnitude of all STFT frames up to time $t$, the network predicts the log-magnitude at time $t+1$. We performed an evaluation of our model on the TIMIT speech dataset (\url{https://catalog.ldc.upenn.edu/LDC93S1}) using the pre-processing and train/test split per~\cite{wisdom2016full}. We use $3640, 192$ and $400$ utterances for the training, validation and test sets, respectively. The results for this experiment of the mean squared error loss are shown in Tab.~\ref{tab:tmt_res}, where our method and \code{expRNN} beat the other approaches by a large margin.

\begin{table}[!t]
    \caption{Each of the architectures uses $\approx 200$k trainable parameters, and we show the mean squared error results (MSE) on the validation and test sets for each model.}
    \vspace{+0.3cm}
    \centering
    \begin{tabular}{cccc}
        \toprule
        Method & Validation MSE & Test MSE & $\kappa$ \\
        \midrule
        \code{euRNN}~\cite{jing2017tunable}           & $16$      & $15.15$   & $378$ \\
        \code{fcuRNN}~\cite{wisdom2016full}           & $14.96$   & $14.69$   & $256$ \\
        \code{scoRNN}~\cite{helfrich2017orthogonal}   & $7.97$    & $7.36$    & $425$ \\
        \code{expRNN}~\cite{lezcano2019cheap}         & $5.52$    & $5.48$    & $425$ \\ 
        Ours                                            & $6.46$    & $6.15$    & $510$ \\
        \bottomrule 
    \end{tabular} \label{tab:tmt_res}
\end{table}

\subsection{Polyphonic music task}

In this task, the network takes as input temporal sequences of $88$-bit one hot encodings of piano keys where one indicates a key that is pressed at a given time. For example, the following trimmed sequence is taken from the MuseData dataset where the $y$-axis represents time and the $x$-axis corresponds to the different played keys.
\begin{figure} [h]\vspace{+0.3cm}
	\centering
	\begin{overpic}[width=0.45\textwidth]{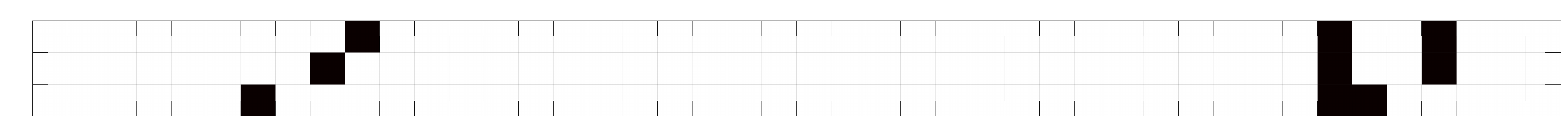} 
	\end{overpic}
	\label{fig:muse_data}
\end{figure}
The goal in this task is to predict the next binary vector for every position in the sequence. 
\begin{table}[!t]
    \caption{In this example we use a similar number of parameters per model ($\approx 350$k), and we report the negative log likelihood loss values obtained for the polyphonic music task on the JSB and MuseData datasets.}
    \vspace{+0.3cm}
    \centering
    \begin{tabular}{cccc}
        \toprule
        Method                                      & JSB dataset & MuseData      \\
        \midrule
        \code{RNN}                                  & $8.77$        & $6.39$        \\
        \code{uRNN}~\cite{arjovsky2016unitary}    & $8.71$        & $6.61$        \\
        \code{euRNN}~\cite{jing2017tunable}       & $8.65$        & $6.52$        \\
        \code{fcuRNN}~\cite{wisdom2016full}       & $8.69$        & $6.66$        \\
        \code{expRNN}~\cite{lezcano2019cheap}     & $8.53$        & $6.53$        \\ 
        \code{nnRNN}~\cite{kerg2019non}           & $8.84$        & $6.2$ &       \\
        Ours                                        & \bm{$8.36$}   & \bm{$6.19$}   \\
        \bottomrule 
    \end{tabular} \label{tab:msc_res}
\end{table}
The loss function for this task is the negative log likelihood. We consider the JSB Chorales dataset which consists of the entire corpus of $382$ four-part harmonized chorales by J. S. Bach, and we also use the MuseData electronic library dataset that includes $783$ orchestral and piano classical music elements. In this task, \code{lvfRNN} obtains significantly better test loss measures in comparison to all other methods for both of the datasets, as we report in Tab.~\ref{tab:msc_res}.

\begin{figure} [!t]
	\centering
	\begin{overpic}[width=.55\linewidth]{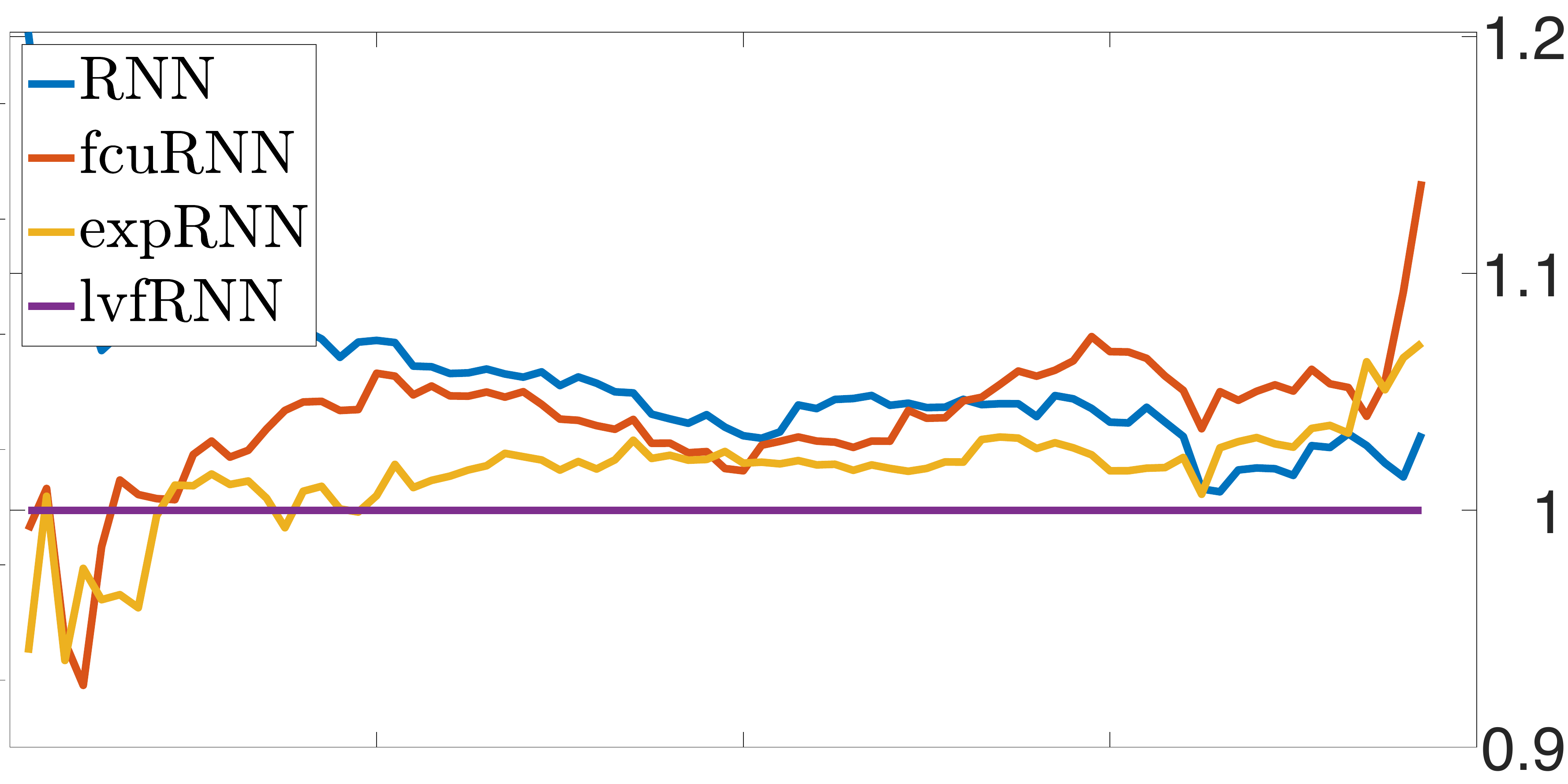} 
	\put(-4,7){\rotatebox{90}{ relative error measure}}
    \put(21,-3){ {chorales examples in the collection}} 
	\end{overpic}\vspace{+0.2cm}
	\caption{We evaluate our trained model on the JSB test data~\cite{allan2005harmonising}, and we show the \emph{relative} negative log likelihood distribution per music element across various techniques.}
	\label{fig:poly_music_eval}
\end{figure}

In order to verify that the better loss measure obtained by our model is also of practical significance, we show in Fig.~\ref{fig:poly_music_eval} the \emph{relative} error measure for a predicted collection obtained with our model in comparison to other recurrent models. 
Namely, the $x$-axis represents the chorales examples in the collection, and the $y$-axis is the error computed~via
$$
\mathrm{d}(Y(i)_{\mathrm{mtd}},\, \tilde{Y}(i)) / \mathrm{d}(Y(i)_{\mathrm{ours}},\, \tilde{Y}(i)) ,
$$ 
where $\mathrm{d}$ is the negative log likelihood function, $\tilde{Y}(i)$ is the true label, $Y(i)_{\mathrm{ours}}$ is the label computed using our method, and $Y(i)_{\mathrm{mtd}}$ is computed using one of the methods in the~legend. Clearly, the \code{lvfRNN} shows a better performance over a wide range of chorales examples.

\section{Conclusion}

In this work, we proposed a novel interpretation of unitary recurrent neural networks using the theory and practice of differential geometry. In particular, orthogonal units are typically designed with respect to their algebraic structure, e.g., the relation between skew-symmetric matrices and the special orthogonal group. Our approach shows that there is an underlying geometrical object associated with every orthogonal model. Namely, we prove that orthogonal neural networks are in fact optimizing in the space of divergence-free vector fields. In practice, our observation allows to exploit tools and results from differential geometry in the context of designing and applying new sequence models; and we suggest a new recurrent model which spans the entire space of vector fields, including those fields with zero divergence. Our method requires the construction of a discrete differential geometry toolbox in latent space, involving the inner product, gradient and divergence operators. We evaluate our model on several common benchmarks, and we show it achieves comparable or better results when compared to recent state-of-the-art baseline~units.

In conclusion, we believe that our work provides a new point-of-view to thinking about recurrent models which is associated with a myriad of theoretical and practical tools for further analysis and design.

\section*{Acknowledgements}
We are grateful for the generous support from Amazon AWS.
N. B. Erichson, and M. W. Mahoney would like to acknowledge the IARPA (contract W911NF20C0035), ARO, NSF, and and ONR via its BRC on RandNLA for providing partial support of this work.
Our conclusions do not necessarily reflect the position or the policy of our sponsors, and no official endorsement should be inferred. M. Ben-Chen acknowledges support from the Israel Science Foundation (grant No. 504/16), and the European Research Council (ERC starting grant no. 714776~OPREP).

\bibliographystyle{unsrt}
\bibliography{lvfrnn_arXiv}

\appendix

\section{Divergence-Free Vector Fields}
\label{app:divfree_skew}

One of the main foundations of our approach is the relation between vector fields and matrices. Here, we prove that divergence-free vector fields in the continuous setting are associated with skew-symmetric directional derivative operators. In what follows, we assume that the functions and vector fields attain the required properties (such as smoothness requirements) such that the following derivations hold.
\setcounter{thm}{0}
\begin{thm}
	The vector field $\mathfrak{v}$ is divergence-free if and only if its operator $\mathcal{D}_\mathfrak{v}$ is skew-symmetric.
\end{thm}

\paragraph{Proof.} Let $\mathcal{M} = \mathbb{R}^m$ be the domain of scalar functions $\mathfrak{f}:\mathcal{M} \rightarrow \mathbb{R}$ and (tangent) vector fields $\mathfrak{v}:\mathcal{M} \rightarrow \mathrm{T}\mathcal{M}$. We denote by $\mathcal{F}$ the space of square integrable scalar functions which vanish at $\pm \infty$, $\mathcal{F} := \{ \mathfrak{f} : \int_\mathcal{M} |\mathfrak{f}(x)|^2 \dd x < \infty, \lim_{x \rightarrow \pm \infty} f(x) = 0\}$, and its associated inner product by $\langle \mathfrak{f}, \mathfrak{g} \rangle_\mathcal{M} := \int_\mathcal{M} \mathfrak{f}(x) \mathfrak{g}(x) \dd x$. It is well-known that $\mathcal{F}$ admits an orthonormal functional basis $\{ \phi_j : \mathcal{M} \rightarrow \mathbb{R} \}_{j=1}^\infty$ such that 
\begin{align}
    \mathfrak{f}(x) = \sum_j \mathfrak{f}_j \phi_j(x) \ , \quad \mathrm{with} \;\; \mathfrak{f}_j := \langle \mathfrak{f}, \phi_j \rangle_\mathcal{M} \ .     
\end{align}
In what follows, we consider the standard differential tools on $\mathcal{M}$ such as the gradient $\nabla$ and the divergence $\nabla(\cdot)$. Let $\mathcal{D}_\mathfrak{v}$ be the operator associated with a divergence-free vector field $\mathfrak{v}$. The operator $\mathcal{D}_\mathfrak{v}$ is defined via $\mathcal{D}_\mathfrak{v} (f)(x) := \langle \mathfrak{v}(x),\, \nabla \mathfrak{f}(x) \rangle$ and thus $\mathcal{D}_\mathfrak{v}$ acts on the space of scalar functions, i.e., $\mathcal{D}_\mathfrak{v} : \mathcal{F} \rightarrow \mathcal{F}$. In the basis $\{ \phi_j \}$, the elements of $\mathcal{D}_\mathfrak{v}$ are given by 
\begin{align*}
    (\mathcal{D}_\mathfrak{v})_{ij} &= \int_\mathcal{M} \phi_i(x) \, \left[ \mathfrak{v}(x) \cdot \nabla \phi_j(x) \right] \dd x \\
    &= - \int_\mathcal{M} \nabla\cdot(\phi_i \, \mathfrak{v} )(x) \, \phi_j(x) \dd x \\
    &= - \int_\mathcal{M} [\mathfrak{v}(x) \cdot \nabla \phi_i(x) + \phi_i(x) \nabla\cdot \mathfrak{v}(x)] \, \phi_j(x) \dd x \\
    &= - \int_\mathcal{M} \left[ \mathfrak{v}(x) \cdot \nabla \phi_i(x) \right] \, \phi_j(x) \dd x \\
    &= - (\mathcal{D}_\mathfrak{v})_{ji} \ ,
\end{align*}
where the second equality holds due to integration by parts on domains with no boundary, the third equality because of vector calculus identities, and the fourth equality uses the zero divergence assumption. Conversely, we note that the above calculation shows that if $\mathcal{D}_\mathfrak{v}$ is skew-symmetric then $\nabla\cdot\mathfrak{v}(x) = 0$ for every $x\in\mathcal{M}$ because its projection onto the basis $\{\phi_j \}_{j=1}^\infty$ is zero for every $j$, i.e., $\langle \phi_j, \nabla\cdot\mathfrak{v} \rangle_\mathcal{M} = 0$.

\section{The Lie Algebra of Vector Field Operators}
\label{app:lie_algebra}

In~\cite{lezcano2019cheap}, the authors parameterize the orthogonal group via the matrix exponential of skew-symmetric matrices. Their approach is advantageous as the matrix exponential and its differential can be computed efficiently. To employ a similar technique on the space of vector fields, we would need that the vector space $\mathrm{VF}_\kappa$ is a Lie algebra. Unfortunately, while it is well-known that vector fields form a Lie algebra with the Lie bracket as a commutator in the continuous setting~\cite{frankel2011geometry}, this property does not carry over to $\mathrm{VF}_\kappa$ as defined in Sec.~\ref{sec:lvfrnn} with the matrix commutator. The key difference is that the nonlinear terms cancel in the continuous setting since the order of differentiation can be arbitrary, whereas skew-symmetric matrices do not commute in general, and thus the nonlinear terms remain in the discrete setup. For instance, 
\begin{align*}
    \mathcal{D}_V = \begin{pmatrix} 
                    0 & 1 & -1 \\
                    -1 & 0 & 1 \\
                    1 & -1 & 0
                    \end{pmatrix} \ , \quad
    \mathcal{D}_U = \begin{pmatrix} 
                    1 & 1 & -2 \\
                    -1 & 0 & 1 \\
                    2 & -1 & -1
                    \end{pmatrix} \ , \quad
    [\mathcal{D}_U, \mathcal{D}_V] = \begin{pmatrix} 
                                      0 & 2 & -2 \\
                                      0 & 0 & 0 \\
                                      -2 & 2 & 0
                                      \end{pmatrix} \ ,
\end{align*}
where $[\mathcal{D}_U, \mathcal{D}_V] = \mathcal{D}_U \mathcal{D}_V - \mathcal{D}_U \mathcal{D}_V$. It is clear that $[\mathcal{D}_U, \mathcal{D}_V]$ is not a directional derivative of a vector field $W \in \mathrm{vf}_\kappa$ as it is not the sum of a skew-symmetric and diagonal matrices $\mathcal{D}_W = R_W - T_W$. We prove the more general result we will need the following lemma.
\begin{lem} \label{lem:commute}
    Let $R$ and $T$ be skew-symmetric and diagonal matrices, respectively. $R$ commutes with $T$ if and only if the following holds for any $i, j$. If $T_{ii} \neq T_{jj}$ then $R_{ij} = 0$.
\end{lem}
\paragraph{Proof.}
The matrices $R$ and $T$ commute if and only if
\begin{align*}
    (T \, R)_{ij} = (T)_{ii} (R)_{ij} = (R)_{ij} (T)_{jj} = (R \, T)_{ij} \ .
\end{align*}
The above relation holds in two cases: 1) $T_{ii} = T_{jj}$, and 2) $T_{ii} \neq T_{jj}$ and $R_{ij} = 0$.

\begin{thm}
    There is no finite-dimension matrix Lie algebra with elements $R + T$ where $R$ is skew-symmetric and $T$ is diagonal under the standard commutator whose skew-symmetric matrices are associated with divergence-free vector fields.
\end{thm}

\paragraph{Proof.} Let $\mathrm{vf}_\kappa$ be the set of vector fields. We consider any construction $\mathrm{VF}_\kappa = \{ \mathcal{A}_V | V \in \mathrm{vf}_\kappa \}$ of directional derivative operators such that $\mathcal{A}_V = R_V + T_V$ with $R_V$ a skew-symmetric matrix and $T_V$ a diagonal matrix. We assume that $\mathrm{VF}_\kappa$ forms a matrix Lie algebra with the standard matrix commutator. Then, the commutator of any two vector fields $U,V \in \mathrm{VF}_\kappa$ reads
\begin{align*}
    [\mathcal{A}_U, \mathcal{A}_V] &= (R_U + T_U)(R_V + T_V) - (R_V + T_V)(R_U + T_U) \\
    &= R_U R_V - R_V R_U + R_U T_V - T_V R_U + T_U R_V - R_V T_U \ . 
\end{align*}

The first term, $R_U R_V - R_V R_U$, is skew-symmetric as can be shown via
\begin{align*}
    R_U R_V - R_V R_U &= R_U^T R_V^T - R_V^T R_U^T = (R_V R_U)^T - (R_U R_V)^T = - (R_U R_V - R_V R_U)^T \ ,
\end{align*}
whereas the other two terms, $R_U T_V - T_V R_U$ and $T_U R_V - R_V T_U$ are symmetric. For instance,
\begin{align*}
    R_U T_V - T_V R_U &= -R_U^T T_V + T_V R_U^T = (R_U T_V)^T - (T_V R_U)^T = (R_U T_V - T_V R_U)^T \ .
\end{align*}
Further, the diagonal part of $R_U T_V - T_V R_U$ (and similarly $T_U R_V - R_V T_U$) is zero, since $R_U$ is skew-symmetric and $T_V$ is diagonal.

The above calculation shows that $[\mathcal{A}_U, \mathcal{A}_V]$ is the sum of a skew symmetric matrix and two symmetric matrices. Denote by $A_{[U,V]} = [A_U,A_V] \in \mathrm{VF}_\kappa$, and let $A_{[U,V]} = R_{[U,V]} + T_{[U,V]}$ be its decomposition into skew-symmetric and diagonal matrices. Then, based on the arguments above, we have that $T_{[U,V]} = 0$, and hence, according to our assumption, $A_{[U,V]}$ represents a divergence-free field and is skew-symmetric. This holds if the symmetric matrix $Z = R_U T_V - T_V R_U + T_U R_V - R_V T_U$ is zero. We note that if both $V$ and $U$ are divergence-free, then $Z=0$. However, not all vector fields in $\mathrm{vf}_\kappa$ have zero divergence. Moreover, if each of the components $R_U T_V - T_V R_U = 0$ and $T_U R_V - R_V T_U = 0$, then we need that $R_U$ (resp. $R_V$) commutes with $T_V$ (resp. $T_U$). In this case, we can apply Lemma~\ref{lem:commute}, and again, not all $W \in \mathrm{vf}_\kappa$ satisfy such a structure. Thus, we assume that both $V$ and $U$ are not divergence-free and the individual components are not zero. Then, to satisfy $Z=0$, we have that 
\begin{align*}
    (R_V)_{ij} ((T_U)_{ii} - (T_U)_{jj}) = (R_U)_{ij} ((T_V)_{ii} - (T_V)_{jj}) \ .
\end{align*}
However, similar to the previous cases, we obtain that $V$ depends on $U$ (and vice versa) to yield a zero $Z$ matrix, but not all vector fields in $\mathrm{vf}_\kappa$ admit the above structure. We conclude that $\mathrm{VF}_\kappa$ is not a Lie algebra as it voids the closure property.

\section{Differential Geometry in Latent Space}
\label{app:diff_geom}

The main challenge in discretizing Eq.~\eqref{eq:vf_op} is that the topology (connectivity) of the latent domain and its differential structure are unknown, unlike triangle meshes where this information is available~\cite{azencot2013operator}. For simplicity, we assume that the underlying $\kappa$-dimensional domain is fully connected with no self-edges. In what follows, we motivate our choices for the operators involved in Eqs.~\eqref{eq:grad}, \eqref{eq:div}, and \eqref{eq:dv_op}. In practice, the choice of the $\cgrad$ operator affects the discretizations of $\cdiv$ and $\mathcal{D}_V$ as well. For the gradient $\cgrad h : \mathbb{R}^{\kappa} \rightarrow \mathbb{R}^{\kappa \times \kappa}$ we simply take the (forward) finite differences between the current node and its neighboring (all other) nodes. Formally, 
\begin{align} \label{eq:grad2}
    (\cgrad h)_{ij} = h_i - h_j \ , \quad i,j = 1,2, ..., \kappa \ .
\end{align}
Now, to construct a divergence operator we may propose an independent discretization. However, it is beneficial that the gradient and divergence satisfy a discrete integration by parts property (see e.g., the proof for Thm~\ref{thm:divfree_skew}). Namely, we require that
\begin{align} \label{eq:int_by_parts}
    \vect(V)^T \vect(\cgrad h) +  (\cdiv V)^T h  = 0 \ ,
\end{align}
for every vector $h \in \mathbb{R}^\kappa$ and vector field $V \in \mathrm{vf}_\kappa$. The combination of Eq.~\eqref{eq:grad2} and Eq.~\eqref{eq:int_by_parts} leads to the following definition of the divergence of a vector field $v$, i.e.,
\begin{align}
    (\cdiv V)_i = \sum_{j=1}^\kappa V_{ji} - V_{ij} \ .
\end{align}
The above definition has a very intuitive interpretation as we sum over the contributions of each of the incoming and outgoing edges related to node $i$. Finally, given $h$ and $V$, we want the discrete $\mathcal{D}_V h$ to respect as many properties as possible of its continuous version, including the relation between divergence-free vector fields and skew-symmetry of the directional derivative. Thus, we arrive at
\begin{equation} \begin{aligned} \label{eq:dir_deriv}
    (\mathcal{D}_V h)_i &= \sum_j (V_{ij} - V_{ji}) (\cgrad h)_{ij} \\ 
    &= \sum_j (V_{ij} - V_{ji}) (h_i - h_j) \\
    &= \left(\sum_{j = 1}^{\kappa} (V_{ij} - V_{ji}) \right) h_i - \sum_{j = 1}^{\kappa} (V_{ij} - V_{ji}) h_j \ .
\end{aligned} \end{equation}
As the above formulation is the same for every $h$, we extract the definition of $\mathcal{D}_V$ and obtain Eq.~\eqref{eq:dv_op}. We emphasize that the diagonal of $V$ does not play a role in $\mathcal{D}_V$, and thus we consider these elements to be zero.

It is straightforward to show that the set $\mathrm{VF}_\kappa = \{ \mathcal{D}_V | V \in \mathrm{vf}_\kappa \}$ forms a vector space with the usual addition of matrices and multiplication by a scalar, the inverse element is $- \mathcal{D}_V = \mathcal{D}_{-V}$, and the identity element is the zero matrix. Moreover, it is immediate from Eq.~\eqref{eq:dir_deriv} that $\mathcal{D}_V$ is a linear operator such that $\mathcal{D}_V \, c = 0$ where $c$ is a constant vector. Unfortunately, elements in $\mathrm{VF}_\kappa$ do not satisfy the Leibniz rule, i.e.,
\begin{align*}
    \left[ \mathcal{D}_V (f \cdot g) \right]_i &= \sum_j (V_{ij}-V_{ji})(f_i g_i - f_j g_j) \\
    &\neq \sum_j (V_{ij}-V_{ji})(2 f_i g_i - f_i g_j - f_j g_i) \\
    &= \left[ \mathcal{D}_V(f) \cdot g + f \cdot \mathcal{D}_V(g) \right]_i \ ,
\end{align*}
for general $f, g \in \mathbb{R}^\kappa$ and $V\in \mathrm{vf}_\kappa$. For instance, if $\kappa=2$, we have that $f=(1,0)^T, g=(0,1)^T$ with $V = [0,\; 1;\; 2,\; 0]$ leads to $\mathcal{D}_V (f \cdot g) = 0$, whereas $\mathcal{D}_V(f)\cdot g + f \cdot \mathcal{D}_V(g) = (1, -1)^T$. Thus, the above discretization~\eqref{eq:dir_deriv} of the directional derivative is not compatible with the continuous case with respect to the Leibniz~rule.


\section{Properties of Directional Derivative Operators}
\label{app:lvfrnn_prop}

\setcounter{prop}{0}
\begin{prop}
    Let $\mathcal{D}_V \in \mathrm{VF}_\kappa$. Then $\mathcal{D}_V$ is normal and has imaginary spectrum if $(\cdiv V) = 0$. 
\end{prop}
\paragraph{Proof.} Let $V \in \mathrm{vf}_\kappa$. It immediately follows that if $(\cdiv V)_i = 0$ for every $i$, then $\mathcal{D}_V$ is skew-symmetric and thus has an imaginary spectrum. Consequently, it is also normal since a real-valued matrix A is normal if $A^T A = A \, A^T$ and we have $\mathcal{D}_V^T \mathcal{D}_V = -\mathcal{D}_V^2 = \mathcal{D}_V \, \mathcal{D}_V^T$. More generally, $\mathcal{D}_V$ is normal if and only if it can be permuted into a block-diagonal matrix with each block having a constant divergence.
\begin{align*} 
    \mathcal{D}_V^T \mathcal{D}_V &= (R_V - T_V)^T (R_V - T_V) = -R_V^2 - T_V \, R_V + R_V \, T_V + T_V^2 \ , \\
    \mathcal{D}_V \, \mathcal{D}_V^T &= (R_V - T_V) (R_V - T_V)^T = -R_V^2 + T_V \, R_V - R_V \, T_V + T_V^2 \ .
\end{align*}
Therefore, it follows that $\mathcal{D}_V$ is normal if $A = -A$ where $A = T_V \, R_V - R_V \, T_V$, which holds when $A=0$, i.e., $T_V$ and $R_V$ commute, and we can apply Lemma~\ref{lem:commute} to obtain the result. 

\begin{prop}
    Let $\mathcal{D}_V \in \mathrm{VF}_\kappa$. Then $\mathcal{C}_V$ is stable if $(\cdiv V)_i \leq 0$ for every node $i=1,2,...,\kappa$.
\end{prop}

\paragraph{Proof.} Let $x \in \mathbb{R}^\kappa$ and $j \in \{ 1,2, ..., \kappa \}$. We assume that $\cdiv(V)_i \leq 0$ for every $i$, and we want to show that the real part of the eigenvalues of $\mathrm{Re}(\lambda_j(\mathcal{C}_V)) \leq 1$. First, we observe that $- x^* \mathcal{D}_V \, x \leq 0$, since
\begin{align*}
    - x^* \mathcal{D}_V x = x^* (T_V - R_V) x = x^* T_V x \leq 0\ ,
\end{align*}
where the second equality holds for any skew-symmetric matrix $R_V$, and the third inequality follows from the non-positive pointwise divergence. Now, let $w_j$ be the eigenvector associated with the eigenvalue $\lambda_j(\mathcal{D}_V)$. It follows that 
\begin{align*}
     \lambda_j(\mathcal{D}_V) |w_j|^2 = w_j^* \lambda_j(\mathcal{D}_V) w_j = w_j^* \mathcal{D}_V w_j \geq 0 \ .
\end{align*}
Therefore, $\mathrm{Re}(\lambda_j(\mathcal{D}_V)) \geq 0$, which is equivalent to the constraint $\mathrm{Re}(\lambda_j(\mathcal{C}_V)) = \mathrm{Re}(1 - \tau \lambda_j(\mathcal{D}_V)) \leq 1$, for any $\tau \ge 0$.

\section{Results for the Copy task} 
\label{app:results}

The copy challenge was first proposed in~\cite{hochreiter1997long} to evaluate the memory capabilities of sequence models. During the training, input and output sequences of length $T+2K$ are randomly generated with respect to an alphabet of size $L$. For instance, an input-output pair for $T=10, K=5, L=9$ could take the following form
\begin{center}
	\begin{tabular}{r l}
		Input: & \texttt{92836-{}-{}-{}-{}-{}-{}-{}-{}-{}-:-{}-{}-{}-} \\ 
		Output: & \texttt{-{}-{}-{}-{}-{}-{}-{}-{}-{}-{}-{}-{}-{}-{}-92836}
	\end{tabular}
\end{center}
Namely, the network goal is to memorize the first $K$ characters and to output them starting at the colon mark while ignoring the in-between $T$ hyphens. The loss function for this task is the cross entropy. Choosing at random the last $K$ characters yields a baseline cross entropy loss of $K \log (L) / (T + 2K)$. Our experiments focus on the setting, $T=200, K=10, L=9$. 

We show the cross entropy loss (CEL) and accuracy results we obtain for this task in Tab.~\ref{tab:cpy_res}. The hidden size is fixed to $128$ units and thus we also list the number of trainable parameters per model. We note that while our approach achieves a loss value which is relatively high, the obtained accuracy is comparable to all other unitary RNNs. Specifically, our model attains $\approx 95\%$, whereas \code{expRNN} yields $100\%$ accuracy. In comparison, a vanilla \code{RNN} fails on the copy task, and it obtains $\approx 16\%$; this result is consistent with previous studies, e.g.,~\cite{arjovsky2016unitary}. 

\begin{table}[!t]
    \caption{We set the hidden state size to $128$ across all models and measure their cross entropy loss and accuracy results. We also list the number of trainable parameters per model.}
    \centering
    \begin{tabular}{cccc}
        \toprule
        Method & CEL & Accuracy & \#params \\
        \midrule
    \code{RNN}                                      & \num{9.5e-2}  & $16\%$ & $19$k \\
        \code{uRNN}~\cite{arjovsky2016unitary}    & \num{3.5e-3}  & $99\%$ & $6.5$k \\
        \code{euRNN}~\cite{jing2017tunable}       & \num{2.1e-3}  & $99\%$ & $18.9$k \\
        \code{fcuRNN}~\cite{wisdom2016full}       & \num{1.6e-3}  & $99\%$ & $10.6$k \\
        \code{expRNN}~\cite{lezcano2019cheap}     & \num{3.5e-6}  & $100\%$ & $10.6$k \\ 
        \code{nnRNN}~\cite{kerg2019non}           & \num{3e-4}    & $100\%$ & $43.6$k \\
        Ours                                        & \num{2.1e-2}  & $95\%$ & $11$k \\
        \bottomrule 
    \end{tabular} \label{tab:cpy_res}
\end{table}

Finally, as this task requires evolution matrices that are (approximately) unitary, we used a midpoint integration rule instead of the explicit Euler step in Eq.~\eqref{eq:euler_step}, namely, given a $V \in \mathrm{vf}_\kappa$ we compute
\begin{align} \label{eq:midpoint_step}
	\mathcal{C}_V &= \left(I + \frac{\tau}{2} \mathcal{D}_V \right)^{-1} \left(I - \frac{\tau}{2} \mathcal{D}_V \right) \ . 
\end{align}
Using an inverse computation during backpropagation naturally increases the computational complexity of our method. In this case, the computational demands are similar to other orthogonal methods such as~\cite{helfrich2017orthogonal, maduranga2019complex}.

\section{Hyperparameters for \code{lvfRNN}}
\label{app:hyperparam}

We detail in Tab.~\ref{tab:params} the various hyperparameters used to train our latent vector field model on the tasks described in Sec.~\ref{sec:results}. We denote by $\kappa$ the size of the hidden layer, $l$ is the number of stacked recurrent layers~\cite{pascanu2013construct}, $\sigma$ is the nonlinearity per Eq.~\eqref{eq:lvfrnn}, opt denotes the optimizer which is Adam~\cite{kingma2014adam} in all cases, LR and LR decay are the learning rate and learning rate decay, respectively, clip and dropout are implemented as described in \cite{pascanu2013difficulty} and \cite{srivastava2014dropout}, respectively, $\tau$ is the time step per Eq.~\eqref{eq:euler_step}, and $\lambda$ is the parameter which balances between the task's loss and the deviation from zero divergence, see Sec.~\ref{sec:imp}.

\begin{table*} [hb]
	\caption{The hyperparameters used for \code{lvfRNN} to obtain the results we reported in the main text.}
	\centering
		\begin{tabular}{lcccccccccc} \toprule
			Task            & $\kappa$  & $l$   & $\sigma$          & opt   & LR                    & LR decay  & clip  & dropout   & $\tau$    & $\lambda$\\
			\midrule
			Copy ($200$)    & $128$     & $1$   & \code{modrelu}    & Adam  & $1\mathrm{e}{-4}$     & $1$       & $-1$  & $0$       & $15$      & $0$ \\
			JSB             & $300$     & $3$   & $\tanh$           & Adam  & $1.5\mathrm{e}{-3}$   & $0.5$     & $15$  & $0.3$     & $1$       & $0$ \\
			MuseData        & $300$     & $3$   & $\tanh$           & Adam  & $1\mathrm{e}{-3}$     & $0.5$     & $20$  & $0.2$     & $3$       & $0$ \\
			PTB ($150$)     & $1400$    & $1$   & $\tanh$           & Adam  & $2\mathrm{e}{-3}$     & $0.5$     & $-1$  & $0$       & $5$       & $0.1$ \\
			TIMIT           & $255$     & $2$   & $\tanh$           & RMSprop & $3\mathrm{e}{-4}$   & $1$       & $15$  & $0$       & $2$       & $0.2$ \\
			\bottomrule 
	\end{tabular} \label{tab:params}
\end{table*}

\end{document}